\documentclass[letterpaper, 10 pt, conference]{ieeeconf}  % Comment this line out if you need a4paper
\usepackage{graphicx}
\usepackage{booktabs}
\usepackage{colortbl}
\usepackage{amsmath}
\usepackage{amssymb}
\usepackage{multirow}
\usepackage{subcaption}
\usepackage{xcolor}
\usepackage{hyperref}

\hypersetup{
    colorlinks=true,
    linkcolor=red
}

\IEEEoverridecommandlockouts                              % This command is only needed if 
\title{\LARGE \bf
LIPM-Guided Reinforcement Learning for Stable and Perceptive Locomotion in Bipedal Robots
}

\author{Haokai Su$^{1}$, Haoxiang Luo$^{1}$, Shunpeng Yang$^{1,2}$, Kaiwen Jiang$^{3}$, Wei Zhang$^{1,5}$ and Hua Chen$^{4,5}$% <-this % stops a space
\thanks{$^{1}$School of Automation and Intelligent Manufacturing (AIM), Southern University of Science and Technology, Shenzhen, China.
        {\tt\small \{12433009, 12232312, 12250045\}@mail.sustech.edu.cn, zhangw3@sustech.edu.cn}}%
\thanks{$^{2}$Department of Civil and Environmental Engineering, The Hong Kong University of Science and Technology, China,
        {\tt\small syangcp@connect.ust.hk}}%
\thanks{$^{3}$Department of Mechanical Engineering, The University of Hong Kong, Hong Kong, China,
        {\tt\small jiangkw\_zz@connect.hku.hk}}%
\thanks{$^{4}$Zhejiang University-University of Illinois Urbana-Champaign Institute (ZJUI), Zhejiang University, Haining, China.
        {\tt\small  huachen@intl.zju.edu.cn}}% 
\thanks{$^{5}$ LimX Dynamics, Shenzhen, China }
}

\begin{document}

\maketitle
\thispagestyle{empty}
\pagestyle{empty}

%%%%%%%%%%%%%%%%%%%%%%%%%%%%%%%%%%%%%%%%%%%%%%%%%%%%%%%%%%%%%%%%%%%%%%%%%%%%%%%%
%%%%%%%%%%%%%%%%%%%%%%%%%%%%%%%%%%%%%%%%%%%%%%%%%%%%%%%%%%%%%%%%%%%%%%%%%%%%%%%%
\begin{abstract}
Achieving stable and robust perceptive locomotion for bipedal robots in unstructured outdoor environments remains a critical challenge due to complex terrain geometry and susceptibility to external disturbances. In this work, we propose a novel reward design inspired by the Linear Inverted Pendulum Model (LIPM) to enable perceptive and stable locomotion in the wild. The LIPM provides theoretical guidance for dynamic balance by regulating the center of mass (CoM) height and the torso orientation. These are key factors for terrain-aware locomotion, as they help ensure a stable viewpoint for the robot's camera. Building on this insight, we design a reward function that promotes balance and dynamic stability while encouraging accurate CoM trajectory tracking. To adaptively trade off between velocity tracking and stability, we leverage the Reward Fusion Module (RFM) approach that prioritizes stability when needed. A double-critic architecture is adopted to separately evaluate stability and locomotion objectives, improving training efficiency and robustness. We validate our approach through extensive experiments on a bipedal robot in both simulation and real-world outdoor environments. The results demonstrate superior terrain adaptability, disturbance rejection, and consistent performance across a wide range of speeds and perceptual conditions.

\end{abstract}

%%%%%%%%%%%%%%%%%%%%%%%%%%%%%%%%%%%%%%%%%%%%%%%%%%%%%%%%%%%%%%%%%%%%%%%%%%%%%%%%
\section{Introduction}
Bipedal robots with their human-like morphology and locomotion patterns have garnered significant attention in the robotics community. However, compared to quadrupedal robots, bipedal systems present more formidable control challenges due to their inherently underactuated dynamics and reduced contact points, making stable locomotion particularly demanding.

In recent years, reinforcement learning has demonstrated remarkable success in addressing legged robot locomotion problems~\cite{joonholee2019scirobotics,siekmann2020learning,siekmann2021blind, kumar2021rma, margolisyang2022rapid,Hwangbo2022vel_est, rudin2022learn_in_minutes}. For quadrupedal robots, these methods have achieved robust and agile locomotion across diverse terrains. In the context of bipedal robots, RL-based approaches have enabled walking control on uneven terrains~\cite{siekmann2020learning, siekmann2021blind}. Integration of visual perception has further enhanced terrain traversability, enabling more complex tasks such as stair climbing, structured obstacle navigation~\cite{duan2024learning, zhuang2024humanoid, long2024learning}, and navigation through risky terrains~\cite{wang2025beamdojo}. However, most of these achievements have been demonstrated in structured laboratory environments. When deployed in outdoor scenarios, bipedal robots face significant challenges, including exteroceptive sensor noise, ground slippage, and unexpected terrain conditions. Consequently, achieving highly traversable and stable perceptive locomotion for bipedal robots remains underexplored.

% Perceptive locomotion in legged robots requires the integration of high-quality exteroceptive and proprioceptive information, allowing the robot to navigate complex environments such as stairs, slopes, and discrete obstacles, both indoors and outdoors. In recent years, reinforcement learning (RL) has made significant strides in enabling perceptive locomotion for legged robots.

% Existing work in terrain understanding and reward design has enabled quadrupedal robots to achieve impressive perceptive locomotion on complex terrains. However, due to the inherently unstable morphology and underactuated nature of bipedal robots, applying these methods to bipedal robots does not always yield the same level of performance. Learning-based approaches for bipedal robots have demonstrated impressive robustness in tasks such as walking, stair navigation, and parkour, but most of these methods have been tested in controlled indoor environments. In outdoor environments, challenges such as exteroceptive sensor noise, slippage, and unexpected ground conditions are inevitable. As a result, achieving stable perceptive locomotion in unstructured outdoor environments for bipedal robots still remains an area with room for improvement.

\begin{figure}[t]
    \centering
    \includegraphics[width=1.0\linewidth]
    {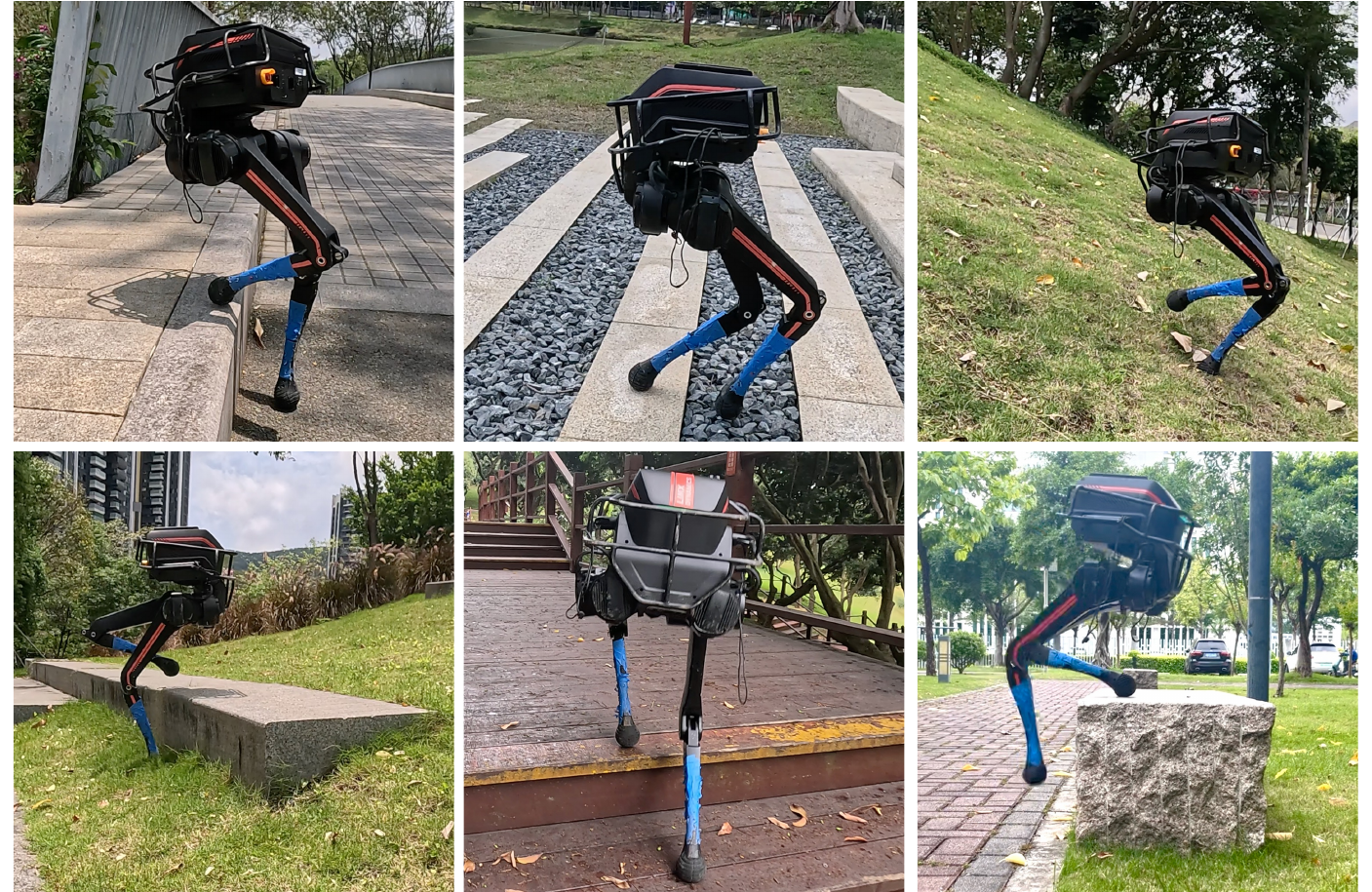} 
    \caption{Demonstration of stable locomotion across diverse and challenging terrains. Our point-foot bipedal robot successfully traverses various outdoor environments—including grass, gravel, slopes, curbs, stairs, and elevated platforms—while maintaining stability without external support or safety tethers.}
    \label{fig:teaser}
    \vspace{-15px}
\end{figure}

In this work, we develop a comprehensive solution for stable perceptive bipedal locomotion. Specifically, we choose a point-foot bipedal robot as the underlying platform, which presents greater control challenges due to its minimal ground contact area and highly underactuated dynamics. To enhance terrain traversability, we first design an efficient vision-based concurrent teacher-student learning framework, which extends the CTS paradigm \cite{wang2024cts} to incorporate perceptive information inputs. Second, given the critical importance of stability in bipedal locomotion, we establish it as a core metric for our robot's movement. We propose a comprehensive bipedal RL reward system that emphasizes stability. Inspired by the well-established LIP model (LIPM), which provides theoretical foundations for stable locomotion, we design a LIPM-guided reward component. Furthermore, rather than requiring the robot to purely track velocity commands, we prioritize stability over velocity tracking through the RFM scheme\cite{jiang2024learning}, ensuring that the robot maintains stable locomotion while achieving desired movement goals. To emphasize the learning of stable rewards during training, we employ a double critic architecture that separately evaluates the LIPM-guided stable reward and locomotion reward, enabling more effective learning of both stability and movement objectives.

As shown in Fig.~\ref{fig:teaser}, the proposed method enables the bipedal robot to achieve stable and reliable locomotion across diverse unstructured outdoor terrains. In summary, our main contributions include:
\begin{itemize}
    \item We propose a novel LIPM-inspired reward design in reinforcement learning of bipedal locomotion. By integrating it with a vision-CTS learning framework, we achieve stable perceptive locomotion for point-foot bipedal robots.
    \item In an effort to balance motion tracking and the stability of biped locomotion, we develop a stability-aware velocity tracking method based on reward fusion. Such an approach decouples command following into direction and magnitude components, which allows for customized prioritization of motion tracking and locomotion stability. 
    \item Through extensive testing in both simulation and real-world environments, we demonstrate that the proposed method enables bipedal robots to achieve stable and perceptive locomotion in outdoor environments. 
\end{itemize}

\section{Related Work}
% \subsection{Teacher-Student Learning for Legged Locomotion}
% Teacher-student learning has emerged as a prominent paradigm in legged robot locomotion, where a teacher policy with access to privileged information guides the learning of a student policy operating with limited sensory input. Traditional two-stage approaches in both blind~\cite{joonholee2020scirobotics,kumar2021rma,wu2023ts_amp} and perceptive locomotion~\cite{Miki2022scirobotics,agarwal2023legged,extremeParkour} first train a teacher policy, followed by student imitation through supervised learning. Recently, Wang et al.~\cite{wang2024cts} introduced the Concurrent Teacher-Student (CTS) framework that trains both policies simultaneously, enabling the student to achieve superior performance through reinforcement learning rather than mere imitation. However, CTS was primarily validated in blind locomotion, limiting its applicability in visually complex environments.
\subsection{Velocity-based vision legged locomotion} 
 Vision-based velocity control for legged locomotion integrates perception and control, enabling legged robots to adapt their movement dynamically based on environmental feedback. Previous methods leverage variational autoencoder(VAE) architectures~\cite{nahrendra2024obstacle, luo2024pie}, teacher-student frameworks~\cite{miki2022learning, agarwal2023legged}, and Transformer models~\cite{luo2024pie, yu2024walking, yang2021learning} to extract compact and task-relevant features from high-dimensional perceptual inputs. These features are subsequently used to condition policy networks, demonstrating excellent performance across diverse terrains on both quadrupedal and bipedal robotic platforms. Notably, \cite{miki2022learning} addresses challenges related to perceptual noise and unreliability, leading to more robust visual-locomotion policies. Despite these advances, existing approaches often struggle in particularly sparse or unstructured terrains, due to limited exploration or conflicting gradients during policy optimization. To mitigate this, \cite{jenelten2024dtc} incorporate trajectory optimization into the RL pipeline to improve foothold selection in sparse environments. However, a common limitation across these works is the prioritization of velocity tracking as the primary control objective. This strict adherence to commanded velocities can compromise balance in scenarios with limited stability margins. While quadrupedal platforms may tolerate such trade-offs, the issue becomes critical for underactuated bipedal robots, especially when navigating gravity-dominated terrains such as downward slopes or stair descents.

\subsection{Bipedal robots locomotion based on LIPM}
Legged robots must constantly stabilize their floating base by interacting with the environment. A common approach to this is the LIPM \cite{kajita20013d,1013335}, which provides closed-form solutions for the CoM and zero moment point (ZMP) dynamics, forming the basis for real-time gait generation and balance control \cite{englsberger2011bipedal}. Extensions of LIPM have been proposed to address the double support phase \cite{li2021trajectory}, improving ZMP transitions and walking trajectory realism. While LIPM has significantly advanced stable control in bipedal locomotion, its simplified dynamics and assumptions can lead to mismatches when applied to real-world scenarios. Factors like uneven terrain, dynamic disturbances, and sensor noise introduce complexities not accounted for by the model, resulting in conservative locomotion behaviors and limiting the robot’s dynamic potential. \cite{castillo2023template} and \cite{lee2024integrating} propose hybrid methods that combine model-based and learning-based approaches to enhance locomotion performance.

\section{Stable CoM Trajectory Generation for Bipedal Robot Locomotion}
\label{sec:CoM Generation}
Point-foot bipedal robots achieve dynamic walking by exploiting gravitational and inertial forces, often through a deliberate forward lean that shifts the CoM beyond the support polygon. This induces a controlled fall that converts gravitational potential energy into forward kinetic energy, enabling efficient locomotion. However, such motion also increases the risk of instability, making the regulation of CoM dynamics and base orientation critical. The LIPM provides a simplified yet powerful framework for reasoning about these dynamics. Under the LIPM assumptions—planar motion of CoM and negligible angular momentum—the evolution of the CoM becomes analytically tractable, enabling predictions about step placement\cite{chen2024contingency}, balance recovery\cite{wieber2006trajectory}, and motion feasibility\cite{li2021trajectory,shi2025bipedal}. Importantly, LIPM serves as the foundation for concepts like capturability\cite{englsberger2011bipedal} and ZMP stability\cite{wieber2006trajectory}, which are widely used in model-based control to ensure safe and stable motion. 

\begin{figure}[htbp]
    \centering
    \includegraphics[width=0.35\textwidth]{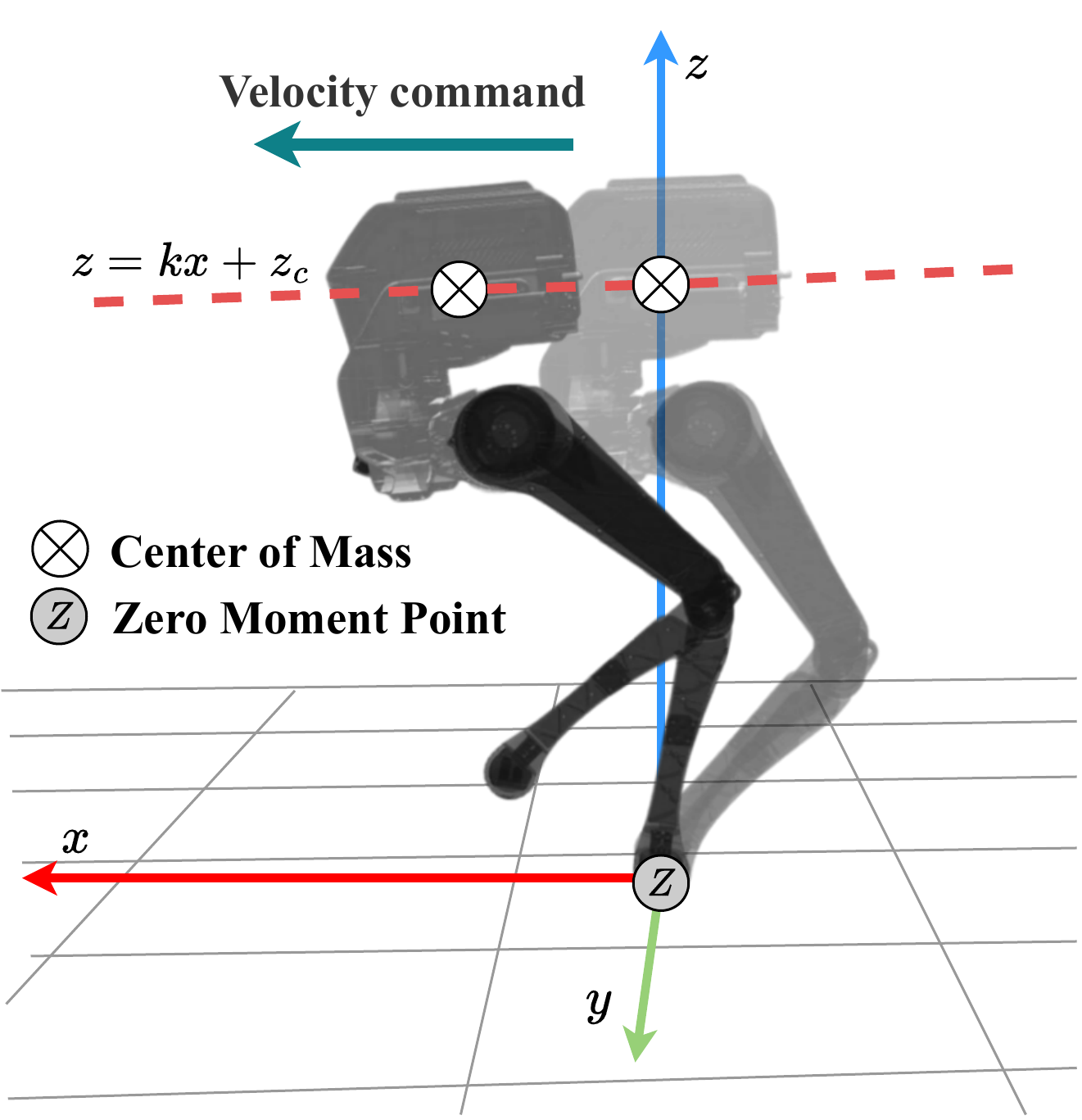}
    \caption{A cross-sectional view of the LIPM model in the $zx$-plane for a point-foot bipedal locomotion system. The motion constraint plane in 3D is reduced to a constraint line in this 2D representation.}
    \label{fig:LIPM System}
\end{figure}

In this section, we leverage the LIPM with a constraint plane to generate stable CoM trajectories. Unlike the traditional constant-height LIPM, our method allows the CoM height to vary dynamically by incorporating a constraint plane\cite{kajita2014kinematics}. As illustrated in Fig.~\ref{fig:LIPM System}, this constraint plane provides a simplified yet effective representation of the robot's dynamics, enabling the robot to adapt to uneven terrains. This flexibility is crucial for achieving stable locomotion in challenging environments, where maintaining a fixed CoM height is often impractical.

According to the dynamics of LIPM, the positions of CoM can be computed as follows:
\begin{equation}
\begin{split}
    \boldsymbol{p}_{\text{com}}=\frac{z}{g}\ddot{\boldsymbol{p}}_{\text{com}}+\boldsymbol{p}_{\text{ZMP}}
\end{split}
\end{equation}
where $z$ denotes the intercept of the centroidal motion constraint plane, $\boldsymbol{p}_{\text{com}}=\left(p_{\text{com},x},p_{\text{com},y}\right)^T$ represents the position of the robot's CoM, $\boldsymbol{p}_{\text{ZMP}}=\left(p_{\text{ZMP},x},p_{\text{ZMP},y}\right)^T$ is the position of the ZMP, and \( g \) is the gravitational acceleration. Inspired by proportional controller, we design the CoM acceleration $\ddot{\boldsymbol{p}}_{\text{com}}$ based on the error of the actual linear velocity $\boldsymbol{v}_{\text{xy}}$ and the linear velocity command $\boldsymbol{v}_{\text{xy}}^{\text{cmd}}$. This allows us to compute the desired CoM position $\hat{\boldsymbol{p}}_{\text{com}}$ as a function of the velocity command:
\begin{equation}
    \hat{\boldsymbol{p}}_{\text{com}} = \boldsymbol{p}_{\text{ZMP}} + \frac{z}{g}k_p(\boldsymbol{v}^{\text{cmd}}_{\text{xy}} - \boldsymbol{v}_{\text{xy}})
    \label{eq:com trajectories}
\end{equation}
where the scalar $k_p$ is feedback gain. This equation not only provides a theoretical foundation for generating desired CoM trajectories but also plays a critical role in the subsequent reward design Section~\ref{sec:reward design} process, where stability and performance are jointly optimized.

\section{Vision-based concurrent teacher-student learning architecture}
\label{sec:vision cts}
% \subsection{Overview}
% We introduce the Vision CTS framework, an extension of the Concurrent Teacher-Student (CTS) framework \cite{wang2024cts}, designed specifically for vision-based legged locomotion. In Vision CTS, agents are organized into two groups: a teacher group and a student group, with the division based on access to observation information. Both groups share the same value and policy networks and are trained concurrently. This concurrent training ensures that the student is optimized directly through reinforcement objectives, rather than relying solely on imitation. By incorporating exteroceptive visual inputs from onboard cameras, Vision CTS enables the student policy to perceive and react to complex terrain. 

\subsection{Overview}
\begin{figure*}[htbp]
    \centering
    \includegraphics[width=1.0\linewidth]{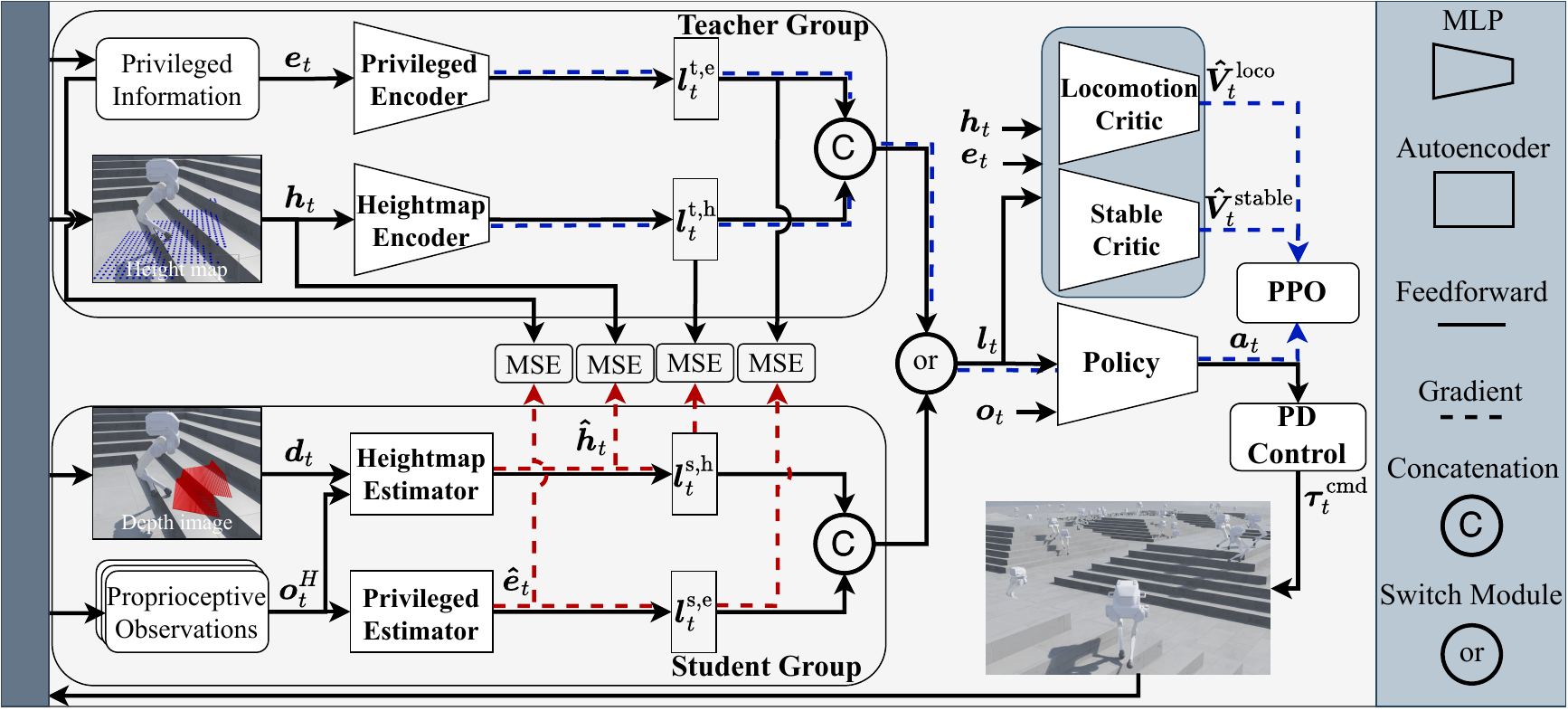}
    \caption{Overview of the Vision-CTS learning framework. Agents are divided into teacher and student groups according to their access to observation modalities. The blue dashed lines indicate PPO gradient \cite{schulman2017proximal} flow, while the red dashed lines denote supervised learning signals.}
    \label{fig:Vision CTS}
\end{figure*}

We build our training framework upon the Concurrent Teacher-Student (CTS) paradigm \cite{wang2024cts}, where both the teacher and student groups share the same value and policy networks and are trained jointly in a single stage. While the original CTS framework was designed for blind locomotion, we extend it by incorporating exteroceptive visual inputs from onboard cameras. We refer to this vision-enhanced framework as Vision-CTS. The overall architecture of our proposed framework is illustrated in Fig.~\ref{fig:Vision CTS}.

% Our approach achieves perceptive locomotion in point-foot bipedal robots through an end-to-end DRL framework that generates the desired joint angle while maintaining stability. As shown in Fig.~\ref{fig:CTS}, our work builds on the Concurrent Teacher-Student (CTS) framework. This framework helps us avoid the problems of two-stage training and uses reinforcement learning to guide the student policy instead of just imitation. Based on these benefits, we apply CTS to our perceptive locomotion system and successfully deploy it on real point-foot bipedal robots.

% \subsection{Composition of Observation}
%  At each timestep, agents in the teacher group has access to the full simulated states of the environment, including the privileged information $\boldsymbol{e}_t$ and the robot surrounding terrain heightmap $\boldsymbol{h}_t$. Agents in the student group has access proprioceptive information $ \boldsymbol{o}_t $, as well as the egocentric depth image $ \boldsymbol{d}_t $. The detailed observations for both groups are summarized in Table~\ref{tab:policy_input}.

\textbf{State Space}: At each timestep, the agents in teacher group has access to the full simulated states of the environment, including the privileged information $\boldsymbol{e}_t$ and the robot surrounding terrain heightmap $\boldsymbol{h}_t$. Agents in the student group has access proprioceptive information $ \boldsymbol{o}_t $, as well as the egocentric depth image $ \boldsymbol{d}_t $. The detailed observations for both groups are summarized in Table~\ref{tab:policy_input}.

\begin{table}[htbp]
    \centering
    \caption{State Spaces for Policy Training}
    \label{tab:policy_input}
    \begin{tabular}{cc|cc}
    \toprule[1.1pt]
    \textbf{Teacher Obs.} & \textbf{Dim} & \textbf{Student Obs.} &\textbf{Dim} \\ 
    \hline
    \rowcolor[gray]{0.8} \multicolumn{2}{c|}{$\boldsymbol{e}_t$} & \multicolumn{2}{c}{$\boldsymbol{o}_t$} \\
    Proprioception GT   & 27 & Base angular velocity & 3 \\
    Base linear velocity   & 3 & Base orientation & 3 \\
    Joint torque   & 6 & Joint positions & 6 \\
    Joint acceleration   & 6 & Joint velocities & 6 \\
    Feet contact force   & 6 & Velocity command & 3 \\
    External force  & 3 & Last action & 6 \\
    \rowcolor[gray]{0.8} \multicolumn{2}{c|}{$\boldsymbol{h}_t$} & \multicolumn{2}{c}{$\boldsymbol{d}_t$} \\
    Heightmap & 441 & Depth image & $80\!\times\!60$ \\
    \bottomrule[1.1pt]
    \end{tabular}
\end{table}

\textbf{Action Space:} The policy outputs a 6-dimensional action vector $\boldsymbol{a}_t$, representing joint position offsets added to the nominal pose. These target positions are tracked by PD controllers with fixed gains ($K_p = 40.0$, $K_d = 2.5$), which generate the torque commands $\boldsymbol{\tau}^{\text{cmd}}_t$.

\subsection{Student Group's Estimators}
The student group's precedent networks comprise two estimators: the privileged estimator and the heightmap estimator. Both are designed with an encoder-decoder architecture and output not only the explicitly estimated information, but also the corresponding latent representations.

\subsubsection{Privileged Estimator}
The privileged estimator employs multilayer perceptrons (MLPs) as both its encoder and decoder. The encoder processes the historical sequence of proprioceptive observations $\boldsymbol{o}^H_t$ and encodes them into a latent representation $\boldsymbol{l}^{\text{s},\text{e}}_t$. The decoder reconstructs the estimated privileged information $\boldsymbol{\hat{e}}_t$ from this latent representation.

\subsubsection{Heightmap Estimator}
The encoder of the heightmap estimator is inspired by the fusion architecture proposed in \cite{agarwal2023legged} for combining visual and proprioceptive information. An MLP extracts features from the historical sequence of proprioceptive observations $\boldsymbol{o}^H_t$, while a convolutional neural network (CNN) processes the current depth image $\boldsymbol{d}_t$. These proprioceptive and exteroceptive features are concatenated and passed through a gated recurrent unit (GRU) to capture temporal dependencies, resulting in the latent representation $\boldsymbol{l}^{\text{s},\text{h}}_t$. This latent vector is subsequently decoded by an MLP to estimate the reconstructed heightmap $\boldsymbol{\hat{h}}_t$.

\subsubsection{Training Objective}
For the student group estimators, the training objective consists of two main components: aligning the latent representations of the student encoders with those of the corresponding teacher encoders, and ensuring that the decoded outputs closely match the ground-truth values. This dual objective is implemented via a composite reconstruction loss function:
\begin{equation}
\begin{aligned} 
\mathcal{L}^{\text {rec}} = &\text{MSE}(\boldsymbol{l}^{\text{s},\text{e}}_t, \boldsymbol{l}^{\text{t},\text{e}}_t) + \text{MSE}(\boldsymbol{l}^{\text{s},\text{h}}_t, \boldsymbol{l}^{\text{t},\text{h}}_t) \\ & + \text{MSE}(\boldsymbol{\hat{h}}_t, \boldsymbol{h}_t) + \text{MSE}(\boldsymbol{\hat{e}}_t, \boldsymbol{e}_t) 
\end{aligned}
\end{equation}

% The policy network is implemented as a MLP that takes the current proprioceptive observations $\boldsymbol{o}_t$ and the latent representations $\boldsymbol{l}_t$ from both the teacher and student groups as inputs. The policy network outputs a 6-dimensional action vector which is used to compute desired joint angles, then translated into joint torque commands $\boldsymbol{\tau}^{\text{cmd}}_t$ via a PD controller. 

% The value network employs a multi-critic architecture with two components: a standard critic for standard rewards and a stable critic dedicated to the stability reward defined in Section~\ref{sec:stable reward}. Both critics take as input the privileged information $\boldsymbol{e}_t$, the heightmap 
% $\boldsymbol{h}_t$ , and the aggregated latent representation $\boldsymbol{l}_t$. 

\subsection{Reward Design}
\label{sec:reward design}
Designing an effective reward function is critical for achieving stable and efficient bipedal locomotion, particularly in unstructured and dynamically challenging environments. This section presents a novel reward framework that explicitly prioritizes stability while preserving accurate velocity tracking. Central to our approach are three key components: (i) \textbf{Stable Reward}, which embeds theoretical insights into balance and CoM control; (ii) \textbf{Stable-Tracking Reward Fusion} that adaptively balances multiple objectives; and (iii) \textbf{Decoupled Velocity Tracking Reward} that separates direction and magnitude to improve training stability. Together, these components overcome the limitations of conventional weighted-sum reward designs by establishing a clear and learnable hierarchy between stability and motion objectives, leading to more robust and adaptable locomotion policies.
% where $\boldsymbol{r}^{\text{loco}}_{\text{reg}}$ regularize the robot to retain smooth and feasible movement, $\boldsymbol{r}_{\text{vel}}$ is the velocity tracking reward, which evaluates how well the robot tracks both linear and angular velocity commands.
\\

\noindent\textbf{Stable Reward:} First, to ensure the robot tracks the CoM trajectories discussed in Section~\ref{sec:CoM Generation}, it is necessary to maintain consistency with the assumptions of the LIPM using a constraint plane. Specifically, we constrain the intercept of the CoM motion constraint plane, $z_{\text{c}}$, to match the robot's upright standing height. However, the normal vector of the constraint plane is left unconstrained, allowing the control policy to dynamically adjust the plane's normal based on external sensory inputs. This design enables the robot to adapt to varying terrain conditions while maintaining stability. Additionally, zero angular momentum around the CoM is enforced to align with the fundamental assumptions of the LIPM. Based on these considerations, the stable reward function $\boldsymbol{r}_{\text{stable}}$ is formulated as follows:
\begin{align}
\label{eq:stable reward}
\boldsymbol{r}_{\text{stable}} &= \exp\left( -\|\hat{\boldsymbol{p}}_{\text{com}} - \boldsymbol{p}_{\text{com}}\|_2 - |z_c - z| \right. \nonumber \\
& \qquad \left. - (|\dot{\theta}_{\text{roll}}| + |\dot{\theta}_{\text{pitch}}|) \right) \nonumber \\
&=\exp\left( -\|\boldsymbol{p}_{\text{e}}\|_2 - |z_{\text{e}}| - \|\boldsymbol{\omega}_{\text{e}}\|_1 \right)
\end{align}
where $\boldsymbol{p}_{e} = \hat{\boldsymbol{p}}_{\text{com}} - \boldsymbol{p}_{\text{com}}$ denotes the CoM tracking error vector, $z_e = z_c - z$ represents the intercept error, and $\boldsymbol{\omega}_e = \left[\dot{\theta}_{\text{roll}}, \dot{\theta}_{\text{pitch}}\right]^T$ is the vector of roll and pitch angular velocities.
\\

\noindent\textbf{Stable-Tracking Reward Fusion:} For the perceptive locomotion challenge, when the bipedal robot loses stability during locomotion, it should slow down or stop to adjust its CoM position and torso orientation, which could be evaluated by (\ref{eq:stable reward}), then resume velocity tracking. This highlights a fundamental priority: stability must take precedence over velocity tracking. However, traditional weighted-sum reward approaches fail to maintain this priority hierarchy effectively. Inspired by the RFM, a method that establishes task priorities through non-linear fusion of reward components, we apply this mechanism to combine stable rewards $r_{\text{stable}} \in (0,1]$ and velocity tracking rewards $r_{\text{vel}} \in (0,1]$:
\begin{equation}
\label{eq:RFM}
    \boldsymbol{r}_t = \boldsymbol{r}_{\text{stable}} + \boldsymbol{r}_{\text{stable}}\boldsymbol{r}_{\text{vel}}
\end{equation}
The partial derivatives of $\boldsymbol{r}_t$ with respect to $\boldsymbol{r}_{\text{stable}}$ and $\boldsymbol{r}_{\text{vel}}$ are computed as shown in (\ref{eq:RFM derivative}):
\begin{equation}
\label{eq:RFM derivative}
    \frac{\partial \boldsymbol{r}}{\partial\boldsymbol{r}_{\text{stable}}} = 1 + \boldsymbol{r}_{\text{vel}}, \quad\frac{\partial \boldsymbol{r}}{\partial\boldsymbol{r}_{\text{vel}}} =  \boldsymbol{r}_{\text{stable}}
\end{equation}
$\boldsymbol{r}_{\text{vel}}$ only takes effect when $\boldsymbol{r}_{\text{stable}}$ is sufficient, meanwhile $\boldsymbol{r}_{\text{stable}}$ always have impact on $\boldsymbol{r}_t$. This design enables the robot to learn behaviors that prioritize maintaining stability before attempting to track velocity commands.
\\

\noindent\textbf{Decoupled Velocity Tracking Reward:} The velocity tracking $\boldsymbol{r}_{\text{vel}}$ is commonly formulated as:
\begin{equation}
    \boldsymbol{r}_{\text{vel}} = \exp{(-\alpha\|\boldsymbol{v}^{\text{cmd}} - \boldsymbol{v}\|_2^2)}
    \label{eq:vel reward}
\end{equation}
where $\alpha$ is a scaling factor, $\boldsymbol{v}^{\text{cmd}}$ represents the desired velocity, and $\boldsymbol{v}$ is the current velocity. This formulation creates a hyperspherical error space around the desired velocity, encouraging the policy to minimize the error and align the robot's actual velocity with the desired value. While effective in prior work, this approach introduces challenges when coupled with stability rewards. Specifically, when the stable reward $\boldsymbol{r}_{\text{stable}}$ is low, the gradient signal for optimizing $\boldsymbol{r}_{\text{vel}}$ becomes weak, leading to degraded velocity tracking performance. 

To address this issue, we decouple linear velocity tracking into two components: direction tracking and magnitude tracking. For direction tracking, we use cosine similarity $\mathcal{D}$ between the desired linear velocity $\boldsymbol{v}_{\text{xy}}^{\text{cmd}}$ and the actual linear velocity $\boldsymbol{v}_{\text{xy}}$:
\begin{equation}
\mathcal{D}(\boldsymbol{v}_{\text{xy}}^{\text{cmd}}, \boldsymbol{v}_{\text{xy}}) = \frac{\boldsymbol{v}_{\text{xy}}^{\text{cmd}} \cdot \boldsymbol{v}_{\text{xy}}}{\|\boldsymbol{v}_{\text{xy}}^{\text{cmd}}\|_2 \|\boldsymbol{v}_{\text{xy}}\|_2}
\end{equation}
For magnitude tracking, we compute the $\ell_2$-norm difference between the magnitudes of the desired and actual velocities. The linear velocity tracking reward is then reformulated as:
\begin{equation}
\label{eq:decouple}
\begin{split}
\boldsymbol{r}^{\text{loco}}_{\text{lin}} &= \exp{(\mathcal{D}(\boldsymbol{v}_{\text{xy}}^{\text{cmd}}, \boldsymbol{v}_{\text{xy}}) - 1)} + \exp{(-\|\|\boldsymbol{v}_{\text{xy}}^{\text{cmd}}\|_2 - \|\boldsymbol{v}_{\text{xy}}\|_2\|^2)} \\
&= \exp{(d_{\text{e}})} + \exp{(m_{\text{e}})}
\end{split}
\end{equation}
Here, $d_{\text{e}}$ represents the directional error, and $m_{\text{e}}$ represents the magnitude error. This decoupled formulation ensures that stability primarily modulates speed magnitude while maintaining directional control, allowing the robot to follow velocity commands even under low-stability conditions.

Therefore, the overall reward function, as described in (\ref{eq:overall reward}), helps the robot achieve stable and efficient locomotion, with details of its three components provided in Table~\ref{tab:reward}.
\begin{equation}
\label{eq:overall reward}
    \boldsymbol{r} = \boldsymbol{r}_{\text{stable}} + \boldsymbol{r}_{\text{stable}}\boldsymbol{r}^{\text{loco}}_{\text{lin}} + \boldsymbol{r}^{\text{loco}}_{\text{reg}}
\end{equation}

\subsection{Double Critic for Stable Reward Learning}
As detailed in Section~\ref{sec:reward design}, the reward design is categorized into two groups: the \textbf{locomotion group} ($r_{\text{lin}}^{\text{loco}}$, $r_{\text{reg}}^{\text{loco}}$) and the \textbf{stability group} ($r_{\text{stable}}$). Following the approach in \cite{zargarbashi2024robotkeyframing}, a double-critic architecture is utilized, with each critic corresponding to one of these reward groups. The \textbf{locomotion critic} is responsible for evaluating the expected return of the locomotion rewards, producing an estimated value of $\boldsymbol{\hat{V}}_t^{\text{loco}}$. Concurrently, the \textbf{stability critic} assesses the expected return of the stability reward, yielding $\boldsymbol{\hat{V}}_t^{\text{stable}}$. The inputs to both critics include the privileged information $\boldsymbol{e}_t$, the heightmap $\boldsymbol{h}_t$, and the aggregated latent representation $\boldsymbol{l}_t$.

\begin{table}[t]
    \centering
    \caption{Reward Terms}
    \label{tab:reward}
    \begin{tabular}{lll}
    \toprule[1.1pt]
    Reward Term & Equation &  Weight \\
    \hline 
    \rowcolor[gray]{0.8} \multicolumn{3}{c}{\textbf{Stable Reward} $r_{\text{stable}}$} \\
    Stable & $\exp{(\boldsymbol{p}_e+z_e+\|\boldsymbol{\omega}_e\|_1)}$ & 1.0 \\
    \hline 
    \rowcolor[gray]{0.8} \multicolumn{3}{c}{\textbf{Tracking Linear Velocity Reward} $r^{\text{loco}}_{\text{lin}}$} \\
    Lin. direction tracking & $\exp{(4d_e)}$ & $0.5$ \\
    Lin. magn. tracking & $\exp{(4m_e)}$ & $0.5 \cdot r_{\text{stable}}$ \\
    \hline 
    \rowcolor[gray]{0.8} \multicolumn{3}{c}{\textbf{Locomotion Reward} $r^{\text{loco}}_{\text{reg}}$} \\
    Ang. velocity tracking  & $\exp\left( -4({\omega}_{z}^\text{cmd} - {\omega}_{z})^2 \right)$ &  $0.5$ \\
    Lin. velocity ($z$)  & ${v}_{z}^2$ &  $-2.0$ \\
    % Orientation  & \Vert$\boldsymbol{g}_{xy}\Vert_2 $ &  $-0.05$ \\
    Joint acceleration & $\ddot{\boldsymbol{q}}^2$ &  $-2.5 \times 10^{-7}$ \\
    Joint power  & $|\boldsymbol{\tau}| |\dot{\boldsymbol{q}}|^T$ &  $-2 \times 10^{-5}$ \\
    Joint Torque & $\Vert\boldsymbol{\tau}\Vert_2^2$ &  $-0.0001$ \\
    Action rate & $\Vert\boldsymbol{a}_t - \boldsymbol{a}_{t-1}\Vert_2^2$ &  $-0.01$ \\
    Action smoothness & $\Vert\boldsymbol{a}_t - 2\boldsymbol{a}_{t-1} - \boldsymbol{a}_{t-2}\Vert_2^2$ &  $-0.01$ \\
    Collision  & $ n_\text{collision} $ &  $-1$ \\
    Joint limit & $ n_\text{limitation} $ &  $-2$ \\
    \bottomrule[1.1pt]
    \end{tabular}
    \vspace{-10px}
\end{table}

\section{Experiment and Results}
\subsection{Training and Deployment}\label{subsec:experiment setup}
% \noindent\textbf{Training and Real Robot.} The RL policy is trained in the Isaac 
%  Sim simulator using massively parallel environments. We applied curriculum learning and a decaying entropy coefficient during training to enhance results and improve the smoothness of the policy on convergence. Domain randomization was used to reduce the gap between sim-to-sim and sim-to-real. We trained the agent in 2048 environments, completing 10,000 iterations, with each iteration consisting of 24 environment steps. After less than 12 hours of training on an NVIDIA RTX 4090, we deployed the policy in Gazebo and on a real robot, both performing inference at 50 Hz. On the real robot, an Intel NUC acts as the onboard computer. It is equipped with an Intel RealSense D435i depth camera, which provides depth images at 30 Hz and crops them to 80 $\times$ 60 before transmitting them to the policy.
We train our locomotion policy using the framework described in Section~\ref{sec:vision cts} and the reward system detailed in Section~\ref{sec:reward design}. Training is conducted in Isaac Lab~\cite{mittal2023orbit} with 2048 parallel environments and a 1:1 ratio between teacher and student groups. A terrain curriculum similar to~\cite{rudin2022learn_in_minutes} was adopted. To enhance the policy's robustness and facilitate effective sim-to-real transfer, we apply extensive domain randomization during training. The specific parameters and their randomization ranges are summarized in Table~\ref{tab:domin rand}. Training was completed in approximately 12 hours on a single Nvidia RTX 4090 GPU. After training, the student policy was deployed on the LimX Dynamic TRON1. All image pre-processing and neural network inference were executed onboard using an Intel NUC. Depth images were acquired from a front-mounted Intel RealSense D435i camera. The policy was updated at 50 Hz, while depth image processing was performed asynchronously at 30 Hz via a multi-threaded pipeline.

\begin{table}[t]
    \centering
    \caption{Domain Randomization}
    \label{tab:domin rand}
    \begin{tabular}{lll}
    \toprule[1.1pt]
    Parameter & Randomization Range &  Unit \\ 
    \hline 
    Payload mass  & $[-1, 3]$ & Kg \\
    Center of mass shift  & $[-3,3]\times[-2,2]\times[-3,3]$ & cm \\
    Friction coefficient & $[0.4, 1.2]$ & - \\
    Restitution coefficient & $[0.25, 0.75]$ & - \\
    Joint $K_p$  & $[0.8, 1.2]\times \text{nominal value}$ & N$\cdot$rad \\
    Joint $K_d$  & $[0.8, 1.2]\times \text{nominal value}$ & N$\cdot$rad/s \\
    Motor Strength $K_d$  & $[0.8, 1.2]\times \text{nominal value}$ & N \\
    % Joint Offset  & $[-0.1, 0.1]$ & rad \\
    System delay & $[0, 20]$ & ms \\
    Camera position & [-10, 10] & mm \\
    Camera pitch angel & [-1, 1] & deg \\
    Camera FOV & [-1, 1] & deg \\
    \bottomrule[1.1pt] 
    \end{tabular}
    % \vspace{-15px}
\end{table}

% \subsection{Training Evaluation}
% We evaluate the terrain adaptation capability of our Vision-CTS training framework by comparing it with the original CTS method~\cite{wang2024cts} and a vision-based two-phase teacher-student learning approach~\cite{extremeParkour} (referred to as Vision-TS below). The main distinction between Vision-CTS and Vision-TS lies in the training paradigm: Vision-TS first trains the teacher policy, and then trains the student policy to imitate the teacher's actions in second phase. All other training configurations are kept identical across the compared methods.

% As shown in Fig.~\ref{fig:terrain_level_comparison}, after training converges, the teacher models from all methods achieve comparable terrain levels. However, the student policy trained with Vision-CTS significantly outperforms those from the other two approaches, achieving approximately 22\% higher terrain levels than the CTS student and about 7\% higher than the Vision-TS student. These results indicate that equipping the student with visual perception in Vision-CTS substantially improves its terrain adaptation capability. Furthermore, the superior performance of the Vision-CTS student demonstrates the advantage of reinforcement learning with direct environment interaction, as opposed to solely imitating the teacher in a two-phase paradigm.

% \begin{figure}[htbp]
%     \centering
%     \includegraphics[width=0.45\textwidth]{figures/terrain_level_comparison.png}
%     \caption{Terrain level comparison during training}
%     \label{fig:terrain_level_comparison}
% \end{figure}

\subsection{Ablation Study}
\label{sec:ablation}
\begin{figure}[htbp]
    \centering
    \begin{subfigure}[b]{0.23\textwidth}
        \centering
        \includegraphics[width=\textwidth]{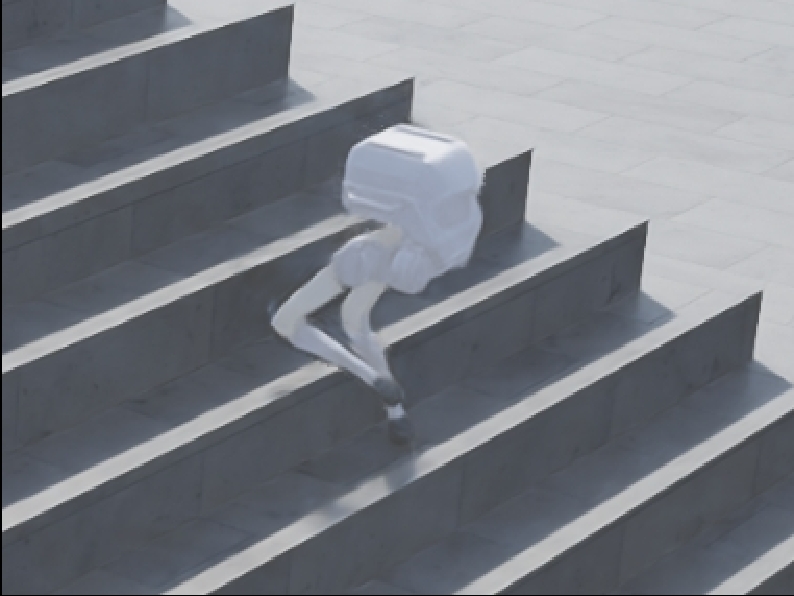}
        \caption{Stairs}
    \end{subfigure}
    \begin{subfigure}[b]{0.23\textwidth}
        \centering
        \includegraphics[width=\textwidth]{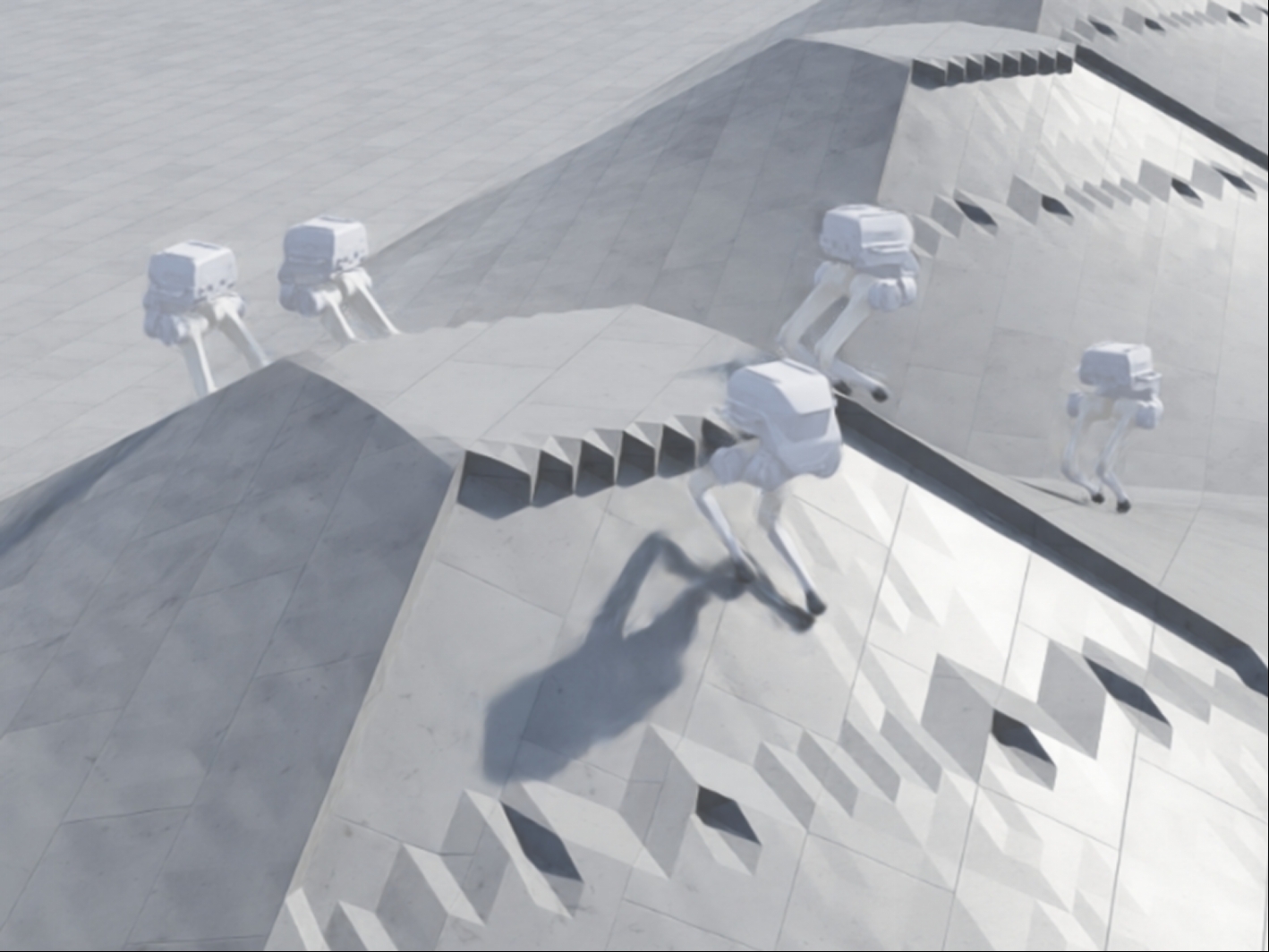}
        \caption{Slopes}
    \end{subfigure}
    \begin{subfigure}[b]{0.23\textwidth}
        \centering
        \includegraphics[width=\textwidth]{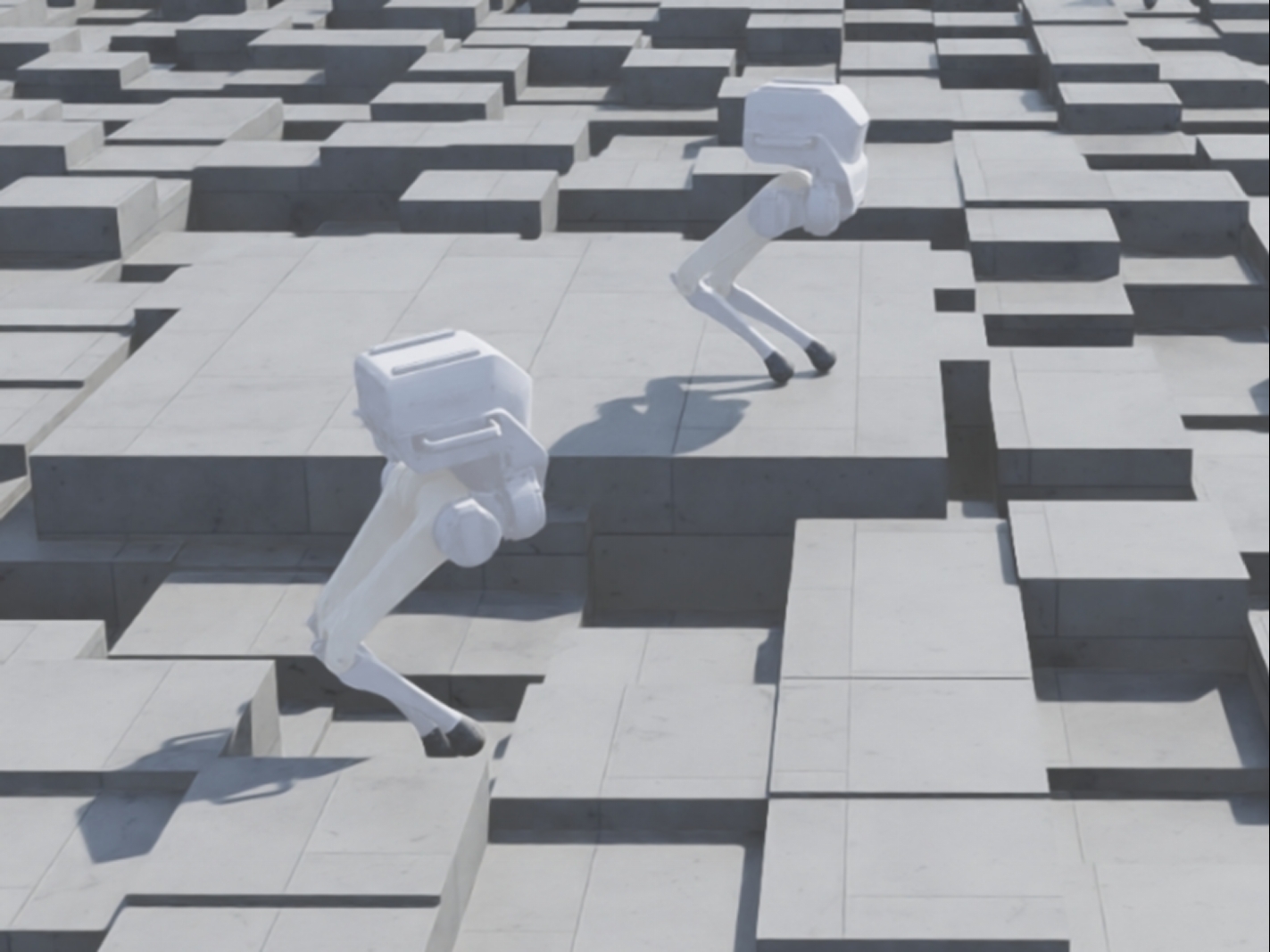}
        \caption{Discrete}
    \end{subfigure}
    \begin{subfigure}[b]{0.23\textwidth}
        \centering
        \includegraphics[width=\textwidth]{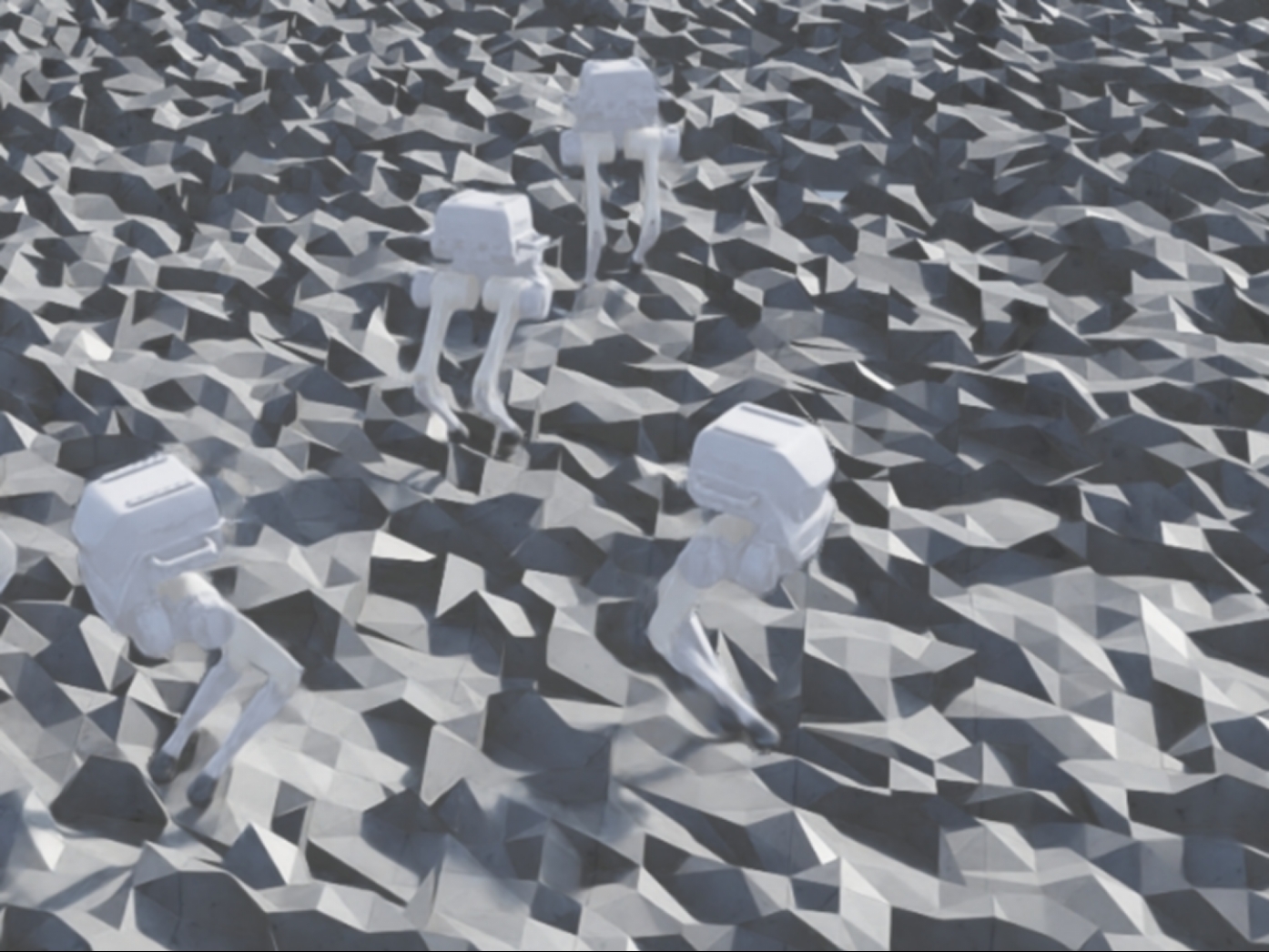}
        \caption{Rough}
    \end{subfigure}\\[0.1cm]
    \caption{Terrains used in the simulation evaluation. The slopes have gradients up to $32.10^{\circ}$. The rough terrain contains uniform noise with heights ranging from 2 cm to 15 cm. The stairs have a height of 20 cm. The discrete obstacles vary in height from 2 cm to 36 cm.}
    \label{fig:terrains}
\end{figure}

We validate our approach in simulation by comparing our method to the following ablations:
\begin{enumerate}
    % \item \textbf{PIE}\cite{luo2024pie}: This one-stage end-to-end learning-based perceptive parkour framework.
    \item \textbf{Ours w/o Stable Reward:}\label{item:ours_wo_stablereward} This is an ablation that removes the stable reward.
    \item \textbf{Ours w/o Stable Critic:}\label{item:ours_wo_stablecritic} This is an ablation that uses a single critic to handle all of the reward.
    \item \textbf{Ours w/o RFM:}\label{item:ours_w_o_rfm} This is an ablation that removes the RFM modules, meaning that no priority is given between velocity tracking and stability keeping.
\end{enumerate}
To ensure fairness, all methods are trained using the same three random seeds and identical training budgets. We then conduct a series of comparative experiments by evaluating the trained policies on the terrains shown in Fig.~\ref{fig:terrains} to assess the effectiveness of our method.

\noindent\textbf{Locomotion Performance in different Terrain.} We evaluate the robot's performance across four terrains shown in Fig. \ref{fig:terrains} using five key metrics. To highlight the contribution of the stable reward to locomotion robustness, we compare our method with the \ref{item:ours_wo_stablereward}) and \ref{item:ours_wo_stablecritic}).
\begin{itemize}
    \item \textbf{Success Rate}: The survival rate of the robot across different terrains.
    \item \textbf{Orientation Error}: 
    The error between the normalized gravity projection and the standard vector $[0,0,-1]$.
    \item \textbf{Angular Velocity Error}: The average angular velocity error in the $x$ and $y$ axes.
    \item \textbf{Velocity Tracking Error}: The average error between desired velocity and actual velocity in body frame.
    % \item \textbf{CoM Tracking}:
    % The tracking error between the desired and actual CoM position.
    \item \textbf{Rec Error}: The error in reconstructing the height map.
\end{itemize}

\begin{table*}[htbp]
\centering
\caption{Performance Comparison Across Terrains. All policies are trained with three different random seeds and evaluated across 1024 distinct environments. Each evaluation is conducted over 500 simulation steps, corresponding to 10 seconds of real-world time. The results are obtained under the most challenging terrain during training, highlighting the robustness and adaptability of the proposed method.}
\label{tab:baslineResults}
\resizebox{\textwidth}{!}{%
\begin{tabular}{lcccccc}
\toprule
\multirow{2}{*}{\textbf{Obstacle}} & \multirow{2}{*}{\textbf{Method}}
& \textbf{Success Rate}$\uparrow$ & \textbf{Orientation Error}$\downarrow$ & \textbf{Angular Velocity Error}$\downarrow$ & \textbf{Velocity Tracking Error}$\downarrow$ & \textbf{Rec Error}$\downarrow$  \\
&  & (\%) & (m/$s^2$) & (rads/s) & (m/s) & (cm) \\
\midrule
\multirow{3}{*}{\textbf{Stairs}}
% & PIE & 62.34 $\pm$ 3.59 & 0.14 $\pm$ 0.02 & 0.91 $\pm$ 0.01 & 0.29 $\pm$ 0.01 & 0.62 $\pm$ 0.06 \\ 
& Ours w/o Stable Reward & 41.43 $\pm$ 5.14 & 0.12 $\pm$ 0.03 & 0.85 $\pm$ 0.03 & 0.29 $\pm$ 0.01 & 2.99 $\pm$ 0.09 \\
& Ours w/o Stable Critic & 52.72 $\pm$ 4.65 & 0.06 $\pm$ 0.01 & 0.7 $\pm$ 0.01 & 0.27 $\pm$ 0.01 & 2.8 $\pm$ 0.11\\
% & Ours w/o RFM  & 47.82 $\pm$ 9.84 & 0.06 $\pm$ 0.01 & 0.65 $\pm$ 0.05 & 0.28 $\pm$ 0.03 & 2.75 $\pm$ 0.17 \\
& Ours  & \textbf{80.30 $\pm$ 4.43} & \textbf{0.06 $\pm$ 0.01} & \textbf{0.64 $\pm$ 0.05} & \textbf{0.26 $\pm$ 0.01} & \textbf{2.57 $\pm$ 0.14} \\
\midrule
\multirow{3}{*}{\textbf{Slope}}
% & PIE & 56.25 $\pm$ 3.62 & 0.14 $\pm$ 0.02 & 1.01 $\pm$ 0.02 & 0.34 $\pm$ 0.02 & 0.65 $\pm$ 0.07\\
& Ours w/o Stable Reward & 66.05 $\pm$ 2.65 & 0.11 $\pm$ 0.03 & 0.91 $\pm$ 0.08 & 0.34 $\pm$ 0.07 & 8.19 $\pm$ 4.38 \\
& Ours w/o Stable Critic & 69.76 $\pm$ 0.56 & 0.08 $\pm$ 0.02 & 0.86 $\pm$ 0.14 & 0.33 $\pm$ 0.07 & 7.85 $\pm$ 4.23\\
% & Ours w/o RFM  & 71.94 $\pm$ 4.84 & 0.06 $\pm$ 0.01 & 0.79 $\pm$ 0.06 & 0.32 $\pm$ 0.04 & 7.47 $\pm$ 3.17 \\
& Ours & \textbf{75.2 $\pm$ 6.24} & \textbf{0.06 $\pm$ 0.01} & \textbf{0.73 $\pm$ 0.11} & \textbf{0.31 $\pm$ 0.04} & \textbf{6.58 $\pm$ 3.81} \\
\midrule
\multirow{3}{*}{\textbf{Discrete}}
% & PIE &52.64 $\pm$ 1.27 & 0.13 $\pm$ 0.02 & 0.9 $\pm$ 0.02 & 0.26 $\pm$ 0.01 & 0.56 $\pm$ 0.04 \\
& Ours w/o Stable Reward & 76.56 $\pm$ 4.91 & 0.11 $\pm$ 0.04 & 0.81 $\pm$ 0.04 & 0.25 $\pm$ 0.02 & 3.64 $\pm$ 0.08 \\
& Ours w/o Stable Critic & 79.26 $\pm$ 4.44 & 0.06 $\pm$ 0.01 & 0.71 $\pm$ 0.01 & 0.24 $\pm$ 0.01 & 3.37 $\pm$ 0.02 \\
% & Ours w/o RFM  & \textbf{84.02 $\pm$ 4.19} & 0.06 $\pm$ 0.01 & 0.66 $\pm$ 0.04 & 0.25 $\pm$ 0.02 & 3.34 $\pm$ 0.13 \\
& Ours  & \textbf{82.81 $\pm$ 3.1} & \textbf{0.05 $\pm$ 0.01} & \textbf{0.64 $\pm$ 0.04} & \textbf{0.24 $\pm$ 0.01} & \textbf{3.25 $\pm$ 0.07} \\
\midrule
\multirow{3}{*}{\textbf{Rough}}
% & PIE & 82.13 $\pm$ 2.9  & 0.12 $\pm$ 0.02 & 0.96 $\pm$ 0.01 & 0.28 $\pm$ 0.01 & 0.55 $\pm$ 0.09 \\
& Ours w/o Stable Reward & 78.94 $\pm$ 1.23 & 0.1 $\pm$ 0.04 & 0.81 $\pm$ 0.03 & 0.26 $\pm$ 0.01 & 2.44 $\pm$ 0.04 \\
& Ours w/o Stable Critic & 81.71 $\pm$ 2.31 & 0.06 $\pm$ 0.0 & 0.68 $\pm$ 0.04 & \textbf{0.24 $\pm$ 0.01} & 2.37 $\pm$ 0.03 \\
% & Ours w/o RFM  & \textbf{85.71} $\pm$ \textbf{3.1} & 0.06 $\pm$ 0.01 & 0.67 $\pm$ 0.03 & 0.25 $\pm$ 0.01 & 2.38 $\pm$ 0.01 \\
& Ours  & \textbf{85.03 $\pm$ 1.29} & \textbf{0.05 $\pm$ 0.01} & \textbf{0.64 $\pm$ 0.01} & 0.26 $\pm$ 0.01 & \textbf{2.33 $\pm$ 0.08} \\
\bottomrule
\end{tabular}%
}
\end{table*}

As shown in Table~\ref{tab:baslineResults}, our method achieves the highest success rates across all terrains, consistently outperforming both ablations. The proposed stability reward, combined with the LIPM-guided CoM tracking objective, effectively reduces the orientation error and the angular velocity error. Improved stability also enables more accurate height-map reconstruction, enhancing the robot’s ability to perceive complex terrains and maintain robust locomotion. This indicates improved dynamic balance and posture control, enabling the robot to maintain robust locomotion under diverse and challenging terrain conditions.
% In contrast, existing perceptive locomotion methods \cite{luo2024pie} still struggle with gravity-dominant terrains, mainly due to the lack of explicit CoM (Center of Mass) guidance during policy learning.

% \noindent\textbf{Vision-CTS vs CTS}
% We compare the terrain level of Vision-CTS and CTS during the training process. The main difference between them is that the Vision-CTS student can access perception information through egocentric depth images, while the CTS student is blind to this information. All other training configurations remain identical. As shown in Fig.~\ref{fig:terrain_level_comparison}, we observe that after training converges, the teacher models of both Vision-CTS and CTS achieve similar terrain levels. However, the Vision-CTS student surpasses the CTS student, achieving approximately 22\% higher terrain levels. This suggests that incorporating visual information into the Vision-CTS model enhances the terrain traversal ability. Notably, during training, the terrain level curve of the Vision-CTS student closely aligns with that of the Vision-CTS teacher, indicating that the Vision-CTS student is able to learn the terrain traversal task effectively from the teacher's guidance.

\begin{figure*}[htbp]
    \centering
    \includegraphics[width=1.0\textwidth]{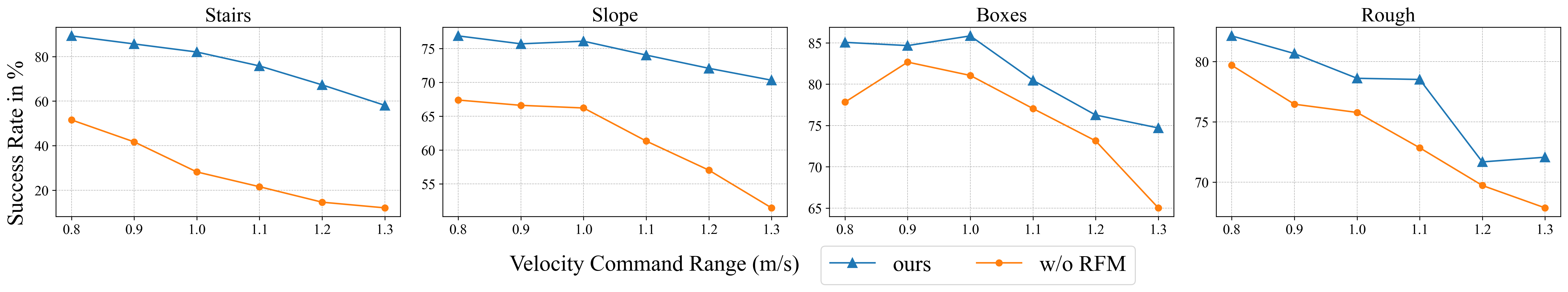}
    \caption{Average Success Rate by velocity dimension for each policy.}
    \label{fig:diffSpeed}
\end{figure*}

\noindent\textbf{Ability at different speeds.}
As shown in Fig.~\ref{fig:diffSpeed}, we compare our method and \ref{item:ours_w_o_rfm}) under different commanded speeds across various terrains. Overall, our method consistently achieves higher success rates across all terrains and velocity commands. By enforcing a control strategy prioritizing stability over velocity tracking, RFM enables the policy to better adapt to high-speed locomotion. This effect is particularly significant on gravity-dominant terrains such as stairs and slope, where maintaining balance becomes more challenging as speed increases.

\noindent\textbf{Push Recovery.}
We evaluated the effectiveness of our method, specifically testing the robot's robustness when subjected to external disturbances. To thoroughly assess the stability of our approach, we conducted experiments under two different conditions: applying moderate disturbances on complex terrains described in Section~\ref{subsec:experiment setup} and extreme disturbances on flat terrain.
\begin{itemize}
    \item Moderate Disturbance Experiment: In this experiment, we applied external wrench within the force range of $[-50, 50]\ N$ and the torque range $[-5, 5]\ N\cdot m$ along the x, y, and z axes.
    \item Extreme Disturbance Experiment: For this experiment, we applied external wrench within the force range of $[-400, 400]\ N$ and the torque range $[-20, 20]\ N\cdot m$ along the x, y, and z axes. 
\end{itemize}
\begin{figure*}[htbp]
    \centering
    \includegraphics[width=0.99\textwidth]{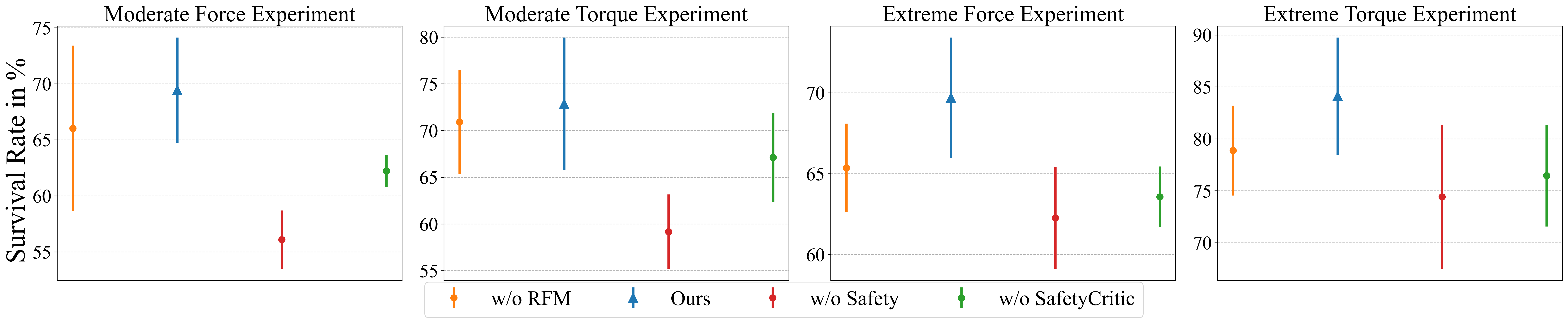}
    \caption{Evaluation of push recovery}
    \label{fig:push recovery}
\end{figure*}
As shown in Fig.~\ref{fig:push recovery}, our method achieves the highest survival rates under both moderate and extreme disturbances across different perturbation scenarios. These results demonstrate the effectiveness of the stability reward in enhancing disturbance recovery.

\begin{table}[h]
\centering
\caption{Absolute Velocity tracking error comparison between the decouple tracking reward (\ref{eq:decouple}) and origin $\ell_2$-norm tracking reward (\ref{eq:vel reward}).}
\label{tab:velrew cmp}
\begin{tabular}{lc}
\toprule
\textbf{Designs} & \textbf{Absoulte Velocity Tracking Error (m/s)$\downarrow$} \\
\midrule
Decoupled Vel Rew & \textbf{0.25 $\pm$ 0.01} \\
$\ell_2$-norm Rew & 0.32 $\pm$ 0.04 \\
\bottomrule
\end{tabular}
\end{table}

\noindent\textbf{Effect of Decoupled Velocity Tracking.}
We evaluate the impact of reward design on velocity tracking performance over composite terrains by comparing the proposed Decoupled Velocity Tracking reward with the conventional $\ell_2$-norm velocity reward. The results in TABLE~\ref{tab:velrew cmp} show that the Decoupled reward formulation mitigates the degradation of velocity tracking performance caused by prioritizing stability through RFM.

% \begin{figure}[htbp]
%     \centering
%     \includegraphics[width=0.4\textwidth]{figures/velTracking.png}
%     \caption{Velocity tracking error comparison between the decouple tracking reward (\ref{eq:decouple}) and origin l2-norm tracking reward (\ref{eq:vel reward}).}
%     \label{fig:velocity tracking}
% \end{figure}

\subsection{Real-world Experiments}
\begin{figure}[htbp]
    \centering
    \includegraphics[width=0.48\textwidth]{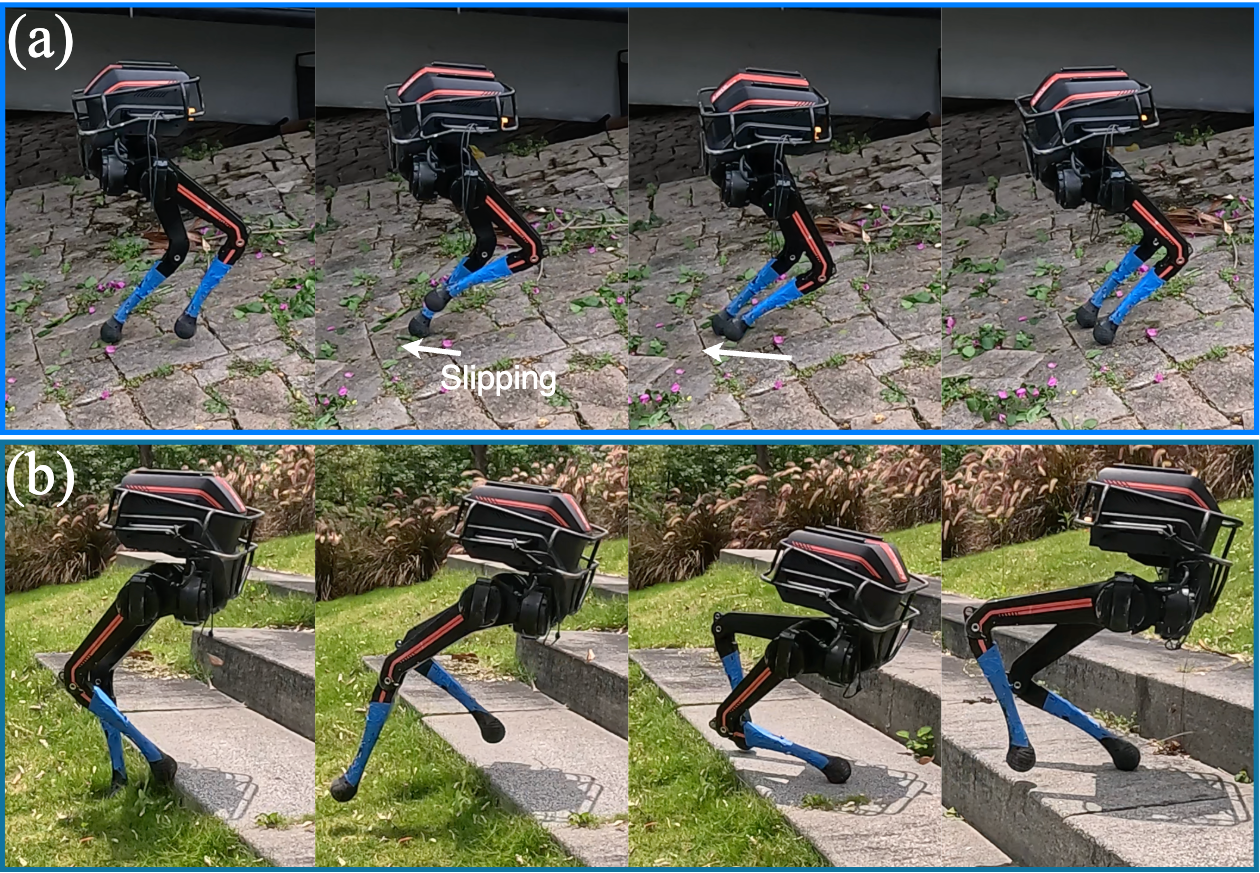}
    \caption{Snapshots of real robot experiment}
    \label{fig:real-robot exp}
\end{figure}

The real-world experiments were conducted across diverse outdoor environments to evaluate our approach. As demonstrated in Fig.~\ref{fig:teaser}, the robot exhibited successful traversal across various challenging terrains, including grass fields, slopes, stairs, gravel paths, and their combinations, demonstrating exceptional adaptability to complex environments. Please refer to our supplementary video for more detailed demonstrations.

The policy exhibited remarkable robustness in handling unexpected disturbances. As illustrated in Fig.~\ref{fig:real-robot exp}(a), when the robot encountered slippery conditions caused by fallen leaves while walking on a wild slope with vegetation, it temporarily suspended velocity tracking and maintained stability before resuming normal locomotion. This adaptive behavior highlights the policy's ability to prioritize stability over motion tracking when necessary.
Fig.~\ref{fig:real-robot exp}(b) demonstrates the robot's resilience in navigating mixed terrain consisting of a grass slope transitioning into stone stairs. At the terrain boundary, dense grass interfered with the depth camera's perception, causing the robot to step on the edge of a stair. Despite this challenging situation, our policy demonstrated exceptional recovery capabilities by quickly adjusting the robot's posture to maintain stability and successfully traversing the subsequent steps.

\section{CONCLUSIONS}

This paper presents a novel approach for achieving stable perceptive locomotion in bipedal robots by integrating LIPM-guided reward design into reinforcement learning. By leveraging the theoretical insights of LIPM, our method enables robots to maintain stability while adapting to diverse and challenging terrains. The proposed RFM ensures that stability is prioritized over velocity tracking when necessary. These innovations collectively address the inherent challenges of bipedal locomotion, such as underactuated dynamics and susceptibility to external disturbances. The significance of our work lies in its ability to bridge the gap between theoretical stability models and practical locomotion strategies, enabling robust and adaptive performance in real-world environments.

Future work will focus on better integrating the dynamic principles revealed by model-based approaches with reinforcement learning frameworks. Specifically, we aim to explore more expressive dynamic models, such as the Spring-Loaded Inverted Pendulum (SLIP) and the Variable-Height Inverted Pendulum (VHIP), which capture a broader range of dynamics, including height variation and viewpoint stability. By leveraging these models, we hope to develop a more comprehensive framework that combines the strengths of model-based insights with the adaptability of learning-based methods. This could enable more dynamic and agile behaviors, enhancing the robot’s ability to traverse unstructured terrains and perform highly dynamic tasks, such as parkour, under a unified perceptive-control framework.

% \addtolength{\textheight}{-1cm}   % This command serves to balance the column lengths

%%%%%%%%%%%%%%%%%%%%%%%%%%%%%%%%%%%%%%%%%%%%%%%%%%%%%%%%%%%%%%%%%%%%%%%%%%%%%%%%

%%%%%%%%%%%%%%%%%%%%%%%%%%%%%%%%%%%%%%%%%%%%%%%%%%%%%%%%%%%%%%%%%%%%%%%%%%%%%%%%



% \section*{ACKNOWLEDGMENT}

% The preferred spelling of the word ÒacknowledgmentÓ in America is without an ÒeÓ after the ÒgÓ. Avoid the stilted expression, ÒOne of us (R. B. G.) thanks . . .Ó  Instead, try ÒR. B. G. thanksÓ. Put sponsor acknowledgments in the unnumbered footnote on the first page.

%%%%%%%%%%%%%%%%%%%%%%%%%%%%%%%%%%%%%%%%%%%%%%%%%%%%%%%%%%%%%%%%%%%%%%%%%%%%%%%%

\bibliographystyle{ieeetr}
\bibliography{reference}

\end{document}